\documentclass[conference]{IEEEtran}
\IEEEoverridecommandlockouts
\usepackage{cite}
\usepackage{amsmath,amssymb,amsfonts}
\usepackage{algorithmic}
\usepackage{graphicx}
\usepackage{textcomp}
\usepackage{xcolor}
\def\BibTeX{{\rm B\kern-.05em{\sc i\kern-.025em b}\kern-.08em
    T\kern-.1667em\lower.7ex\hbox{E}\kern-.125emX}}
\begin{document}

\title{Predicting Compact Phrasal Rewrites with Large Language Models for ASR Post Editing\\
}

\author{\IEEEauthorblockN{Hao Zhang}
\IEEEauthorblockA{\textit{Google Research} \\
haozhang@google.com}
\and
\IEEEauthorblockN{Felix Stahlberg}
\IEEEauthorblockA{\textit{Google Research} \\
fstahlberg@google.com}
\and
\IEEEauthorblockN{Shankar Kumar}
\IEEEauthorblockA{\textit{Google Research} \\
shankarkumar@google.com}
}
\maketitle

\begin{abstract}
Large Language Models (LLMs) excel at rewriting tasks such as
text style transfer 
and grammatical error correction.
While there is considerable overlap between the inputs and outputs in these tasks,
%Although the output in these tasks often significantly overlaps with the input, 
the decoding cost still increases with output length, regardless of the amount of overlap.
%number of overlaps.
By leveraging the overlap between the input and the output, Kaneko and Okazaki \cite{kaneko-okazaki-2023-reducing} proposed model-agnostic edit span representations to compress the rewrites to save computation. They reported an output length reduction rate of nearly 80\% with minimal accuracy impact in four rewriting tasks. In this paper, we propose alternative edit {\em phrase} representations inspired by phrase-based statistical machine translation.
We systematically compare our phrasal representations with their span representations. We apply the LLM rewriting model to the task of Automatic Speech Recognition (ASR) post editing and show that our target-phrase-only edit representation has the best efficiency-accuracy trade-off. On the LibriSpeech test set, 
our method closes 50-60\% of the WER gap between the edit span model and the full rewrite model while losing only 10-20\% of the length reduction rate of the edit span model.

\end{abstract}


\section{Introduction}

Large Language Models pretrained on vast amounts of texts and then 
fine-tuned, instruction-tuned, or prompted
for generation tasks have obtained remarkable success in the past few years \cite{JMLR:v21:20-074, NEURIPS2020_1457c0d6, DBLP:journals/jmlr/ChowdheryNDBMRBCSGSSTMRBTSPRDHPBAI23, anil2023palm2technicalreport, openai2024gpt4technicalreport}.
%One of the key areas such models excel at is text rewriting, paraphrasing and grammatical error correction in particular. 
These models excel at text rewriting tasks, and 
%paraphrasing 
text style transfer
\cite{reif-etal-2022-recipe} and grammatical error correction \cite{rothe-etal-2021-simple, fang2023chatgpthighlyfluentgrammatical} in particular.

However the superior quality of these models comes along with a steep increase in the cost of computation. 
%To support broad deployment and high frequency usage, 
To enable broad deployment for a large user base,
it is crucial to reduce the computational cost while maintaining the accuracy. 

One of the common characteristics of the above-mentioned rewriting tasks is that their output often repeats spans of text in the input. Exploiting the common sub-strings between the input and the output can result in more compact output representations. LLMs can be fine-tuned on examples that map inputs to their compact rewrite representations instead of plain rewrites. At inference time, such a decoding output needs to be composed with the input to expand into complete rewrites.
Kaneko and Okazaki \cite{kaneko-okazaki-2023-reducing} proposed one such representation, which is a numerical span indexing into the input sequence followed by a target phrase that will substitute the source phrase in the given span. We propose two new alternative representations. The first one uses a source-target phrase pair
%pair of source phrase and target phrase
to represent each rewrite pattern, analogous to phrase pairs used by phrase-based statistical machine translation \cite{koehn-etal-2003-statistical}. 
%The second one uses a target phrase solely but covers left and right context words also appearing in the input.
The second representation only uses a target phrase along with left and right context words that appear in the input.
We call the new representations {\em phrase} representations to distinguish them from the {\em span} representation   \cite{kaneko-okazaki-2023-reducing}.

The clear advantage of compact representations over complete rewrites is that the number of decoding steps, and hence the computational cost of inference, is reduced. \cite{kaneko-okazaki-2023-reducing} reported an output length reduction rate of 80\%.
The disadvantage of such representations is that decoding errors can lead to inconsistency
%mismatches 
with the input sequence in the expansion stage, causing an error propagation effect. For example, using the numerical span representation, if the left index or the right index is off by one, the source phrase to be substituted will also be off by one. The concatenation of the context and the substitution can therefore become disfluent. With phrase representations, context words are provided before and after substitution phrases,
which can alleviate the problem of disfluency upon substitution.
%. The problem of disfluency upon substitution can be alleviated.
However, phrase representations have their own problems too.
The predicted context phrase may not match the input, which makes it necessary to discard the subsequent rewrite.
%When the context phrase, which is now to be predicted, does not match the input, the rewrite following it has to be abandoned.
The main focus of the paper is to evaluate the efficiency-accuracy trade-off of the different representations.
%empirical performance of the different representations and recommend which one has the best efficiency-accuracy trade-off.

We choose Automatic Speech Recognition (ASR) post editing as the task for applying LLM-based rewrite models and report word error rates (WER) and output length reduction rates using the span representation and the full rewrite models as the baselines.

Our contributions include the following:
\begin{itemize}
    \item We propose a compact edit string representation with superior efficiency-accuracy trade-off than \cite{kaneko-okazaki-2023-reducing}.
    \item We apply edit representation based rewriting LLMs to the task of ASR output correction. To the best of our knowledge, our work is the first to combine compact rewriting with generative LLMs to achieve substantial ASR WER reduction with manageable decoding cost.
\end{itemize}

\begin{table*}[htb]
  \centering
      \caption{Example output under various representations. We underline the numerical spans or substrings to be matched against the source text in the expansion stage.}

    \begin{tabular}{c|c}
        \hline
         {\em source text} & { Since we do not to bring cash to pay for the transportation fee , enormous time have been saved} \\
         {\em target text} & { Since we do not {\it need} to bring cash to pay for the transportation fee , enormous time {\it has} been saved}\\
         \hline
         \hline
         {\em span} & { \underline{4 4} need, \underline{16 17} has}\\
         \hline
         {\em phrase pair} & { rewrite: \underline{not to}, not need to, rewrite: \underline{time have been}, time has been} \\
         \hline
         {\em target only} & { rewrite: \underline{not} need \underline{to}, rewrite: \underline{time} has \underline{been}}\\
        \hline
        \vspace{0.1in}
    \end{tabular}
    \label{tab:examples}
\end{table*}

\begin{figure}[ht]
%    \centering
    \begin{tabular}{c}
    \includegraphics[width=\columnwidth]{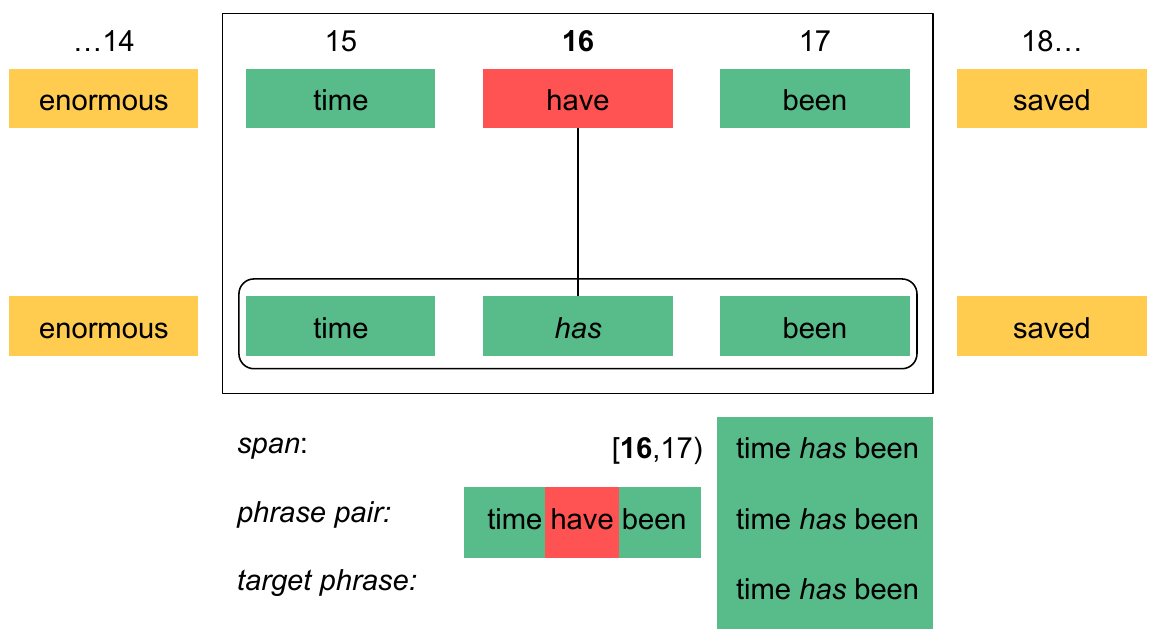}\\    
    \end{tabular}
    \vspace{0.1in}
    \caption{Visualization of the span, phrase pair, and target phrase representations.}    
    \label{fig:phrases}
\end{figure}

\section{Rewrite Representations}\label{sec:compression_expansion}
Mathematically speaking, for rewrite examples $(\mathbf{x}, \mathbf{y})$, where $\mathbf{x}$ is the input string and $\mathbf{y}$ is the output string, there is a compression function $\mathrm{C}$ and an expansion function $\mathrm{E}$ satisfying
\begin{equation}\label{eqn:formula}
\mathrm{E}(\mathbf{x}, \mathrm{C}(\mathbf{x}, \mathbf{y}))=\mathbf{y}
\end{equation}
with the constraint that $|\mathrm{C}(\mathbf{x},\mathbf{y})|$ has a much smaller average value than $|\mathbf{y}|$. At training time, examples are converted to $(\mathbf{x}, \hat{\mathbf{y}}=\mathrm{C}(\mathbf{x}, \mathbf{y}))$, where $\hat{\mathbf{y}}$ is the reference output. At inference time, the final output is obtained by applying the expansion function $\mathrm{E}(\mathbf{x}, \mathbf{y'})$ where $\mathbf{y'}$ is the decoding output for $\mathbf{x}$. The edit representations in this section differ in the choice of the function pair $\mathrm{C}$ and $\mathrm{E}$.

\subsection{Edit Span Representation}\label{sec:span_rep}
The span representation \cite{kaneko-okazaki-2023-reducing} is derived from a word alignment graph $\mathbf{a}$ between $\mathbf{x}$ and $\mathbf{y}$. Given a bipartite alignment graph between the input sequence and the output sequence, we can identify pairs of word spans between the two sides. Each span pair is a local rewrite instance indicating that the source span is substituted by the target span.  In practice, the alignment is derived from the Levenshtein distance algorithm with the guarantee that the alignment links are monotonically ordered. It is always feasible to represent the entire rewrite as a sequence of local rewrite spans.
For LLMs to predict the rewrites, we need a string representation of the span pairs. In their paper, the span representation is 
%in the form of 
specified as
$(i, j,\;\mathbf{y}_{\mathbf{a}(i...j)})$, where $\mathbf{a}(i...j)$ is the corresponding target span of a source span $i...j$ and $\mathbf{y}_{\mathbf{a}(i...j)}$ is the target phrase in this span.
Under this representation, $\mathrm{C}$ is the concatenation of the ordered span representations:
\begin{equation}\label{eqn:span}
 \mathrm{C} = \oplus_{(i,j) \in \mathbf{a}}{(i,j,\;\mathbf{y}_{\mathbf{a}(i...j)})}.
\end{equation}
$\mathrm{E}$ is the program of applying the ordered local rewrites to the input sequence.
\subsection{Phrase Pair Representation}
The representation in Equation~\ref{eqn:span} is concise. It uses a pair of integers to represent a source span. However, this implies that LLMs have to count source tokens and generate indexing integer tokens interleaved with content tokens following the predefined format. The structured representation introduces brittleness to the model. A prediction error in the integer token sub-sequence can have a cascading effect when the entire output rewrite string is parsed and applied to the input.
Instead, we resort to a natural language representation, which uses the source phrase $\mathbf{x}_{i...j}$ for a span $(i, j)$ directly as the prefix for the target phrase $\mathbf{y}_{\mathbf{a}(i...j)}$.
%The challenge is that if the span $(i,j)$ is small, the chance that $\mathbf{x}_{i...j}$ is non-unique increases, making the representation semantically ambiguous. 
However, one downside of our representation is that when the span $(i,j)$ is small, the subsequence $\mathbf{x}_{i...j}$ can be ambiguous, introducing errors into the expansion step.
A solution is to add more context to $\mathbf{x}_{(i...j)}$ as well as $\mathbf{y}_{\mathbf{a}(i...j)}$ to make it much less likely to be ambiguous by extending the phrase pair to both the left and the right.

Formally, the new function $\mathrm{C}$ is
\begin{eqnarray}\label{eqn:source_phrase}
 \mathrm{C} =  \oplus_{(i,j) \in \mathbf{a}}  {(\texttt{W}:\;\mathbf{x}_{(i-k...j+k)},\; \mathbf{y}_{\mathbf{a}(i-k...j+k)})},
\end{eqnarray}
where $k$ is called the {\em dilation span}, $\texttt{W}$ is a natural language prompt word like $\texttt{rewrite}$.
The expansion function $\mathrm{E}$ involves parsing the pattern, %$(\texttt{rewrite}:...,\;...)$ 
prefix string matching, and replacement on the input.

\subsection{Target Phrase Representation}
The representation in Equation~\ref{eqn:source_phrase} using both source phrases and target phrases has the disadvantage of being verbose. Dilation spans on both sides of a phrase, which are intended to make phrases less ambiguous, can exacerbate the problem. However, dilation spans on target phrases can often be sufficient for disambiguation when the span size is three or higher. We can ignore source phrases and just use dilated target phrases as they contain both anchor text in the input and replacement text in the output. The following is the new compression function.

\begin{eqnarray}\label{eqn:target_phrase}
 \mathrm{C} =  \oplus_{(i,j) \in \mathbf{a}}  {(\texttt{W}:\; \mathbf{y}_{\mathbf{a}(i-k...j+k)})}
\end{eqnarray}

String matching and replacement in the implementation of function $\mathrm{E}$ deals with discontiguous dilation spans in the form of $\mathbf{y}_{\mathbf{a}(i-k...i-1)}...\mathbf{y}_{\mathbf{a}(j+1,j+k)}$.

In Table~\ref{tab:examples}, we show actual examples of edit representations. In Fig~\ref{fig:phrases}, we show the alignment of the source phrase ``\underline{time} have  \underline{been}`` and the target phrase ``\underline{time} has \underline{been}`` as well as the three different representations for this edit. In this example, the word ``have`` and the the word ``has`` are aligned to form a phrase pair. We dilate the phrases on both sides to the left and to the right, forming the dilated phrase pair. We also call the words in the dilation regions {\em anchor text} as they can be used to locate the exact position of the target phrase within the source text.
Under the target only representation,  anchor text ``\underline{not} ... \underline{to}'' has two matches (``\underline{not} \underline{to}'' and ``\underline{not} to bring cash \underline{to}''') in the source text. For such cases, we prefer the leftmost and closest pair to break ties. As a result, we can replace ``\underline{not} \underline{to}'' to ``\underline{not} need \underline{to}''.
This example demonstrates that the span representation is the most concise of all three. The two phrase representations are close to concise natural language edit suggestions humans provide. The target only representation is almost as concise as the span representation.

\section{Experiments}
\begin{figure}[ht]
    \centering
    \begin{tabular}{c}
    \includegraphics[width=0.5\columnwidth]{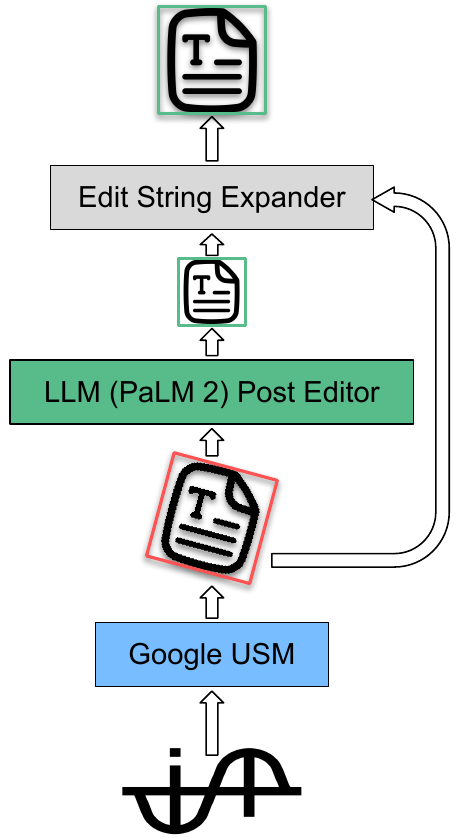}\\    
    \end{tabular}
    \caption{Overview of the entire ASR system with the edit-representation-based LLM post editor.}
    \label{fig:workflow}
\end{figure}

\begin{table*}[htb]
    \centering
    \caption{Results of ASR (USM) post editing models based on PaLM 2 Otter. {\em full} has the lowest WER but has a high computational cost proportional to the average output length. {\em span} \cite{kaneko-okazaki-2023-reducing} is most efficient with the fewest output tokens. {\em target only} closes most of the WER gap between {\em span} and {\em full} while approaching the length reduction rate of {\em span}.  }    
    \begin{tabular}{c|r|r|r|r}
                    & test-clean & & test-other & \\
                    & {\em WER} & {\em Avg Output Length} & {\em WER} & {\em Avg Output Length}  \\
                    \hline
                    \hline
         {\em USM}  &  6.6 & - & 11.4 & \\
         \hline         
         {\em full} &  2.7 (-59\%) & 20 & 6.2 (-46\%) & 18 \\
         {\em span} &  3.4 (-48\%) & 4 (-80\%)  & 7.5 (-34\%) & 5 (-72\%) \\
         \hline
         {\em phrase pair} & 3.0 (-54\%) & 7 (-65\%)  & 7.1 (-38\%) & 10 (-44\%) \\
         {\em target only} & 3.0  (-54\%) & 6 (-70\%) & 6.8 (-40\%)& 8 (-56\%) \\                 
    \end{tabular}
    \vspace{0.1in}

    \label{tab:main}
\end{table*}

\begin{figure*}[htb]
%    \centering
    \begin{tabular}{cc}
    \includegraphics[width=\columnwidth]{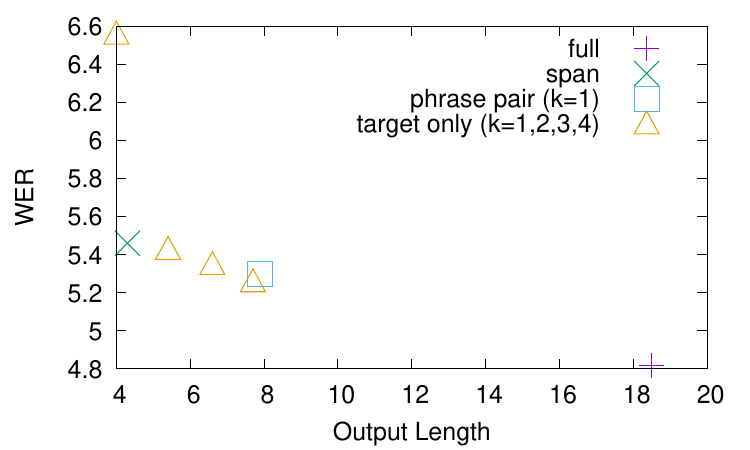}&   
    \includegraphics[width=\columnwidth]{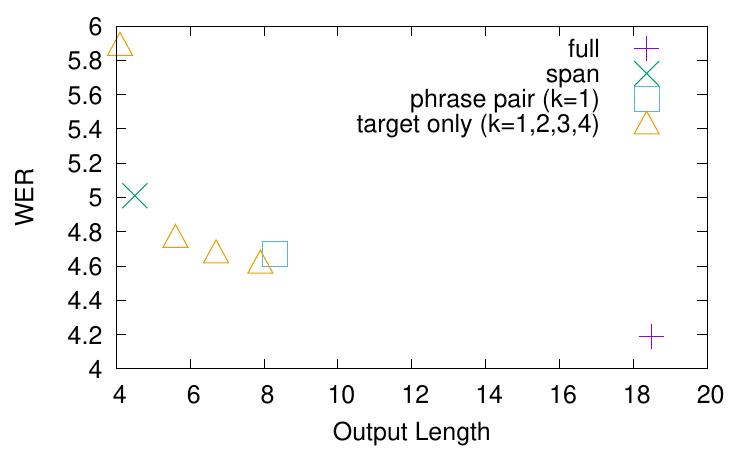}    
    \end{tabular}
    \vspace{0.1in}
    \caption{WER versus output length on the {\em dev} set. Left: PaLM 2 Gecko model. Right: PaLM 2 Otter model. {\em target only} with dilation span size 3 (the third ${\displaystyle \bigtriangleup}$ from the left) is the best strategy. }    
    \label{fig:wer_vs_len}
\end{figure*}

We use a decoder-only LLM for the task of correcting the output of a fast first pass Automatic Speech Recognition (ASR) model. The first pass model is a streaming model that decodes as audio comes in without the full context of the future. Therefore there is sufficient headroom for error correction using a pre-trained LLM with the full ASR transcription as the input. The task can be viewed as a variant of grammatical error correction \cite{brockett-etal-2006-correcting}.

The ASR model we use is the Google USM model \cite{zhang2023googleusmscalingautomatic}. The LLMs we use are the PaLM 2 Gecko and Otter models \cite{anil2023palm2technicalreport}. We fine-tune LLMs on the LibriSpeech \cite{7178964} training set using the {\em dev} set for hyper-parameter and checkpoint selection and the test sets for final comparisons. The ASR model is frozen in our experiments. We fine-tune the entire Transformer LLM model to minimize the cross entropy loss on the transcription reference set given the ASR transcription generated by the frozen USM model as the prefix to the LLM decoder. The entire system is depicted in Figure~\ref{fig:workflow}. We use two baselines. One is {\em full rewrite} model that uses the reference transcription directly. The other is {\em span rewrite} \cite{kaneko-okazaki-2023-reducing} that uses the representation in Section~\ref{sec:span_rep}. We are interested in two metrics. The quality metric is word error rate (WER) after expanding edit representations. The efficiency metric is decoder output length reduction rate.

Table~\ref{tab:main} summarizes the main results. We show that the span representation indeed incurs more accuracy loss than the phrase representations. On the clean test set, {\em target only} is able to close 57\% of the accuracy gap between {\em span} and {\em full}, while losing 12.5\% of the length reduction rate. On the noisier other test set, {\em target only} is able to close 54\% of the accuracy gap, while losing 22.2\% of the length reduction rate.

\subsection{Efficiency and Accuracy Trade-offs}
In Figure~\ref{fig:wer_vs_len}, we plot WER versus output length for two model sizes: Gecko and Otter, and varying values of the phrase dilation hyper-parameter $k$ in Equation~\ref{eqn:source_phrase} and Equation~\ref{eqn:target_phrase}. Both the phrase pair and the target phrase only strategies yield lower WER with slightly longer outputs than the span strategy. Overall, when $k$ is 3, the target phrase only strategy has the best trade-off. The trend is consistent across the two PaLM 2 model sizes.

\subsection{Recovery Rate of Compact Representations}
In Section~\ref{sec:compression_expansion}, we formulated the problem as selecting a pair of compression function $\mathrm{C}$ and expansion function  $\mathrm{E}$ to satisfy Equation~\ref{eqn:formula}. The span representation is exact and unambiguous. So when $\mathrm{E}$ is applied, the 
equality is satisfied for all training examples. The phrase representations can be ambiguous and depend on the dilation spans to reduce the likelihood of multiple matches when the expansion function is applied. Table~\ref{tab:recover} summarizes the recovery rates, which is the percentage of examples in the {\em dev} set that satisfy Equation~\ref{eqn:formula}. The phrase pair representation has sufficient source context in the source phrase so that its recovery rate is very close to 100\%. For the target only representation, word bigrams ($k=2$) or trigrams ($k=3$) surrounding target phrases are sufficient for uniquely identifying their source side counterparts in most cases.

\begin{table}[htb]
    \centering
    \caption{Recovery rate of phrasal representations. {\em phrase pair} (k=1) and {\em target only} (k=3) have a recovery rate close to 100\%.}    
    \begin{tabular}{c|c}
         {\em representation} & {\em recovery rate} \\
         \hline
         {\em phrase pair ($k=1$)} & 99.98\% \\
         \hline
         {\em target only ($k=1$)} & 96.80\% \\
         {\em target only ($k=2$)} & 99.50\% \\
         \hline
         {\em target only ($k=3$)} & 99.80\% \\
         \hline         
         {\em target only ($k=4$)} & 99.80\% \\         
    \end{tabular}

    \label{tab:recover}
\end{table}
%\section*{Acknowledgements}

\section{Related Work}
Orthogonal efforts to speed up decoding include speculative decoding \cite{10.5555/3618408.3619203,chen2023acceleratinglargelanguagemodel}. They leverage the overlap in output distributions between a less accurate faster model and a more accurate slower model as well as hardware accelerators for parallel computing. They do not incur accuracy loss and are not limited to rewriting tasks. Combining compact representations with speculative decoding has the potential for even more speedups.

LLMs have been used for ASR correction in ranking and generation \cite{pu2024multistagelargelanguagemodel}. \cite{leng2021fastcorrect} model edit operations for efficient non-autoregressive decoding. It is possible to perform hybrid decoding with LLMs: predicting which spans need to be rewritten followed by auto-regressive decoding of output rewrites.

\section{Limitations}
Concise edit representations presented in the paper are derived from the Levenshtein distance algorithm. The phrases are not linguistically meaningful or optimal from machine learning point of view. They are only minimal according to the edit distance. Going beyond edit distance to use differentiable functions for compression and expansion is an interesting open area for research.

The dilation spans we use to anchor phrases in the input are applied uniformly and equally on the left and right of each span of interest. It is likely that longer left context is more useful than right context since the decoder progresses from left to right.

%We have not explored different formats for the rewrite phrases. 

With discontiguous phrases, The expansion stage of the target only representation is more algorithmically involved than the other two compact representations.

Finally, we have not experimented with the latest and largest LLMs. %It is possible that prompt engineering is sufficient to let these models generate concise rewrites. 
It is to be seen if the gap between full rewrite and edit representations can be reduced further with very large LLMs.

\section{Conclusions}
We propose two edit phrase representations for rewriting tasks that compactly represent the differences between input and output strings.
%We propose edit phrase representations to compactly represent the differences between the input string and the expected output string. 
We use LLMs to predict such edits and expand the edits into complete rewrites with a deterministic string matching and replacement algorithm. Our work is a further development of the span representation \cite{kaneko-okazaki-2023-reducing}. For the task of ASR post editing, we close 50-60\% of the WER gap between the most efficient model and the most accurate model, while only slowing down decoding by 10-20\% relative to the efficient representation. Future work points to learnable compact representations.

\bibliography{IEEEabrv,IEEEexample.bib}

\end{document}